\documentclass{article}

     \PassOptionsToPackage{numbers, compress}{natbib}
\usepackage[final]{airov24}

\usepackage[utf8]{inputenc} % allow utf-8 input
\usepackage[T1]{fontenc}    % use 8-bit T1 fonts
\usepackage{hyperref}       % hyperlinks
\usepackage{url}            % simple URL typesetting
\usepackage{booktabs}       % professional-quality tables
\usepackage{amsfonts}       % blackboard math symbols
\usepackage{nicefrac}       % compact symbols for 1/2, etc.
\usepackage{microtype}      % microtypography
\usepackage{xcolor}         % colors
\usepackage{multirow} % todo: okay?
\usepackage{amsmath} % todo: okay?
\usepackage{graphicx}  % todo: okay?
\usepackage{subcaption} % todo: okay?

\title{Transparency Techniques for Neural Networks trained on Writer Identification and Writer Verification}

\author{%
  Viktoria Pundy, Marco Peer and Florian Kleber\\
  Computer Vision Lab\\
  TU Wien\\
  Vienna, Austria\\
  \texttt{e1633403@student.tuwien.ac.at}, \texttt{mpeer@cvl.tuwien.ac.at}, \\\texttt{kleber@cvl.tuwien.ac.at} \\
}

\begin{document}

\maketitle

\begin{abstract}
  Neural Networks are the state of the art for many tasks in the computer vision domain, including  Writer Identification (WI) and Writer Verification (WV). The transparency of these ''black box'' systems is important for improvements of performance and reliability. For this work, two transparency techniques are applied to neural networks trained on WI and WV for the first time in this domain. The first technique provides pixel-level sa\-liency maps, while the point-specific saliency maps of the second technique provide information on similarities between two images. The transparency techniques are evaluated using deletion and insertion score metrics. The goal is to support forensic experts with information on similarities in handwritten text and to explore the characteristics selected by a neural network for the identification process. For the qualitative evaluation, the highlights of the maps are compared to the areas forensic experts consider during the identification process. The evaluation results show that the pixel-wise saliency maps outperform the point-specific saliency maps and are suitable for the support of forensic experts.
\end{abstract}

\section{Introduction}
In the last decade, Deep Neural Networks (DNNs) have become a powerful method for various tasks in the computer vision domain and are now a standard for computer vision tasks \cite{samek_2017_evaluating, zhang_2021_survey, zurowietz_2020_interactive, zhang_2017_deep}. This includes WI and WV, where the goal is to identify the author of a handwritten text \cite{schomaker_2008_writer, tang_2016_textindependent}.
The goal of WI is to rank the most similar handwritten documents for a query document and listing the documents of the same author at the top \cite{tang_2016_textindependent, wang_2021_towards}. In contrast, the goal of WV is to determine if two given handwritten texts were written by the same author \cite{schomaker_2008_writer}.  The characteristics of a handwriting depend on the author and environmental circumstances such as the writing tool, geographic location and upbringing of the author \cite{adak_2019_empirical, keglevic_2018_learning, schomaker_2008_writer}, allowing for the identification of a person by their handwriting \cite{bulacu_2007_text-independent}. 

Attention has been drawn to machine learning transparency \cite{samek_2017_evaluating}. Transparency techniques aim to present insight into the internal decision process by displaying the input features a neural network has selected to correctly process the given input \cite{guidotti_2018_survey, samek_2017_evaluating, zhang_2021_survey}. This is for example done to detect artefacts in the training data, which influence the network \cite{samek_2019_towards}, and to support the disclosure of these systems for legal reasons \cite{zhang_2021_survey}.

For this work, transparency techniques are applied to neural networks trained on WI and WV to gain an overview of the expressiveness and validity of the visualizations generated by the selected techniques when using handwritten text images as input. The application on such networks provides information on important areas in the handwritten text and, therefore, expedites the analysis process. Previous work in the domain of transparency techniques mainly focuses on the applicability for networks trained on computer vision tasks, where the input images are rich with information such as colours or multiple objects contained within the image. So far, these techniques have been used to analyse networks trained on tasks including, without limitation, image classification and image retrieval for both natural images \cite{ruderman_1994_statistics} and medical scans \cite{hu_2022_xmir, zheng_2020_towards}, as well as face recognition and person re-identification \cite{zhu_2021_visual}. The use of transparency techniques for networks trained on WI and WV has not yet been a focus of research. This work explores the applicability of these techniques for networks trained on distinguishing different writers. The goal is to use the techniques to provide visual information on the input features selected by the network to successfully execute its task and retrieve information on the internal decision process of the network. The performance of the selected techniques on the WI and WV networks is qualitatively and quantitatively evaluated. Further, the goal is to provide forensic experts with a visualization which supports their decision of an authorship for a handwritten text by highlighting relevant areas.

In summary, this paper provides the following contribution:
\begin{itemize}
    \item We provide a quantitative and qualitative evaluation of the two selected transparency techniques on the three datasets CVL \cite{kleber_2013_cvl}, Firemaker \cite{firemaker} and ICDAR2013 \cite{louloudis_2013_icdar} and show that the pixel-wise saliency maps performs better than the point-specific saliency maps.
    
    \item This research presents the first approach for transparency of WI and WV neural networks. We adapt the existing methodology proposed by Hu et al.~\cite{hu_2022_xmir} for the evaluation of the transparency techniques to accommodate the binarization of the input images by altering the handwriting information, i.e. black pixels, in the input image. This is done to investigate the applicability of the techniques to WI and WV networks.

\end{itemize}

\section{Methodology}
Two transparency techniques are selected for this work. The first technique by Kobs et al. \cite{kobs_2021_different} provides pixel-level saliency maps. The second technique by Zhu et al. \cite{zhu_2021_visual} generates point-specific saliency maps for image pairs. 

\subsection{Pixel-Wise Saliency Maps}
The pixels in the pixel-wise saliency maps are highlighted based on their contribution to the embedding output of the neural network. The saliency map is generated using the difference between an input image $I$ and a base image. The authors propose the use of a black image as base image for the calculation of the maps \cite{kobs_2021_different}. For this methodology, a white image is used as it represents a white sheet of paper without information in the form of handwriting. For the generation of the saliency map, the following gradients are calculated \cite{kobs_2021_different} as $s(\textbf{I}) = \partial d( \textbf{x}_\textbf{I}, \textbf{x}_{base}) / \partial \textbf{I},$
%\begin{equation}
%    s(\textbf{I}) = \frac{\partial d( \textbf{x}_\textbf{I}, \textbf{x}_{base})}{\partial \textbf{I}},
%\end{equation}
where $d$ is a distance function for two embedding vectors. The Cosine Similarity is used as distance function. $\textbf{x}_I$ and $\textbf{x}_{base}$ are the embeddings for image $I$ and the base image, respectively. The generated map is then normalized between 0 and 1, where 1 implies a high contribution and 0 implies no contribution to the network output. In order to reduce the noise of the gradients, $n$ image variants are created by applying a random mask of white pixels to the input image. The resulting gradients are then averaged. This is based on the Smooth-Grad method proposed by the authors \cite{kobs_2021_different}. The value for $n$ is set to 4.

\subsection{Point-Specific Saliency Maps}
The technique for the point-specific saliency maps provides two types of saliency maps. The overall saliency map displays which areas in the input images contribute towards the similarity between the two images and the point-specific saliency map provides information on the similarity of one pixel in one image to all areas in the other image \cite{zhu_2021_visual}.
For a CNN architecture with a GAP and FC layer without bias after the last convolutional layer, the method can be applied as follows \cite{zhu_2021_visual}. The Cosine Similarity is used, which can be rewritten for our network as
\begin{align}
    S = \frac{(\textbf{x}_q \cdot \textbf{x}_r)}{|\textbf{x}_q||\textbf{x}_r|}
    = \frac{\sum\limits_k GAP(\textbf{A}^q_k)GAP(\textbf{A}^r_k)}{|\textbf{x}^q||\textbf{x}^r|} \\
     = \frac{1}{z} \sum_k (\sum_{i,j}\textbf{A}^q_{i,j,k}\sum_{x,y}\textbf{A}^r_{x,y,k})
     = \frac{1}{z} \sum_{i,j,x,y}(\sum_k \textbf{A}^q_{i,j,k}\textbf{A}^r_{x,y,k}),
\end{align}
where $\textbf{A}^q$ and $\textbf{A}^r$ represent the feature maps of the last convolutional layers for a query image $q$ and a retrieval image $r$. $\textbf{x}_q$ and $\textbf{x}_r$ are the embedding outputs for $q$ and $r$, respectively. $(i,j)$ is a point in $q$ and $(x,y)$ is a point in $r$. The point-specific saliency map for the two points $(i,j), (x,y)$ is then defined as $\sum\limits_k \textbf{A}^{q}_{i,j,k}\textbf{A}^r_{x,y,k}$. The overall saliency map for one image is calculated as $\sum\limits_{i,j} (\sum\limits_k \textbf{A}^{q}_{i,j,k}\textbf{A}^r_{x,y,k})$, i.e., a summation over all pixels for one image \cite{zhu_2021_visual}.

\section{Evaluation}
\label{sec:evaluation}
The selected transparency techniques are quantitatively and qualitatively evaluated. First, the setup for the training of the neural networks is described. Then, the metrics used to measure the performance of the techniques and the used datasets are described. Finally, the results of the evaluation are presented.

\subsection{Experiment setup}
This work uses embedding neural networks trained on WI and WV for the evaluation of the transparency techniques. For the WI networks, ResNet18, ResNet20 and ResNet50 are used \cite{wang_2021_towards, rasoulzadeh_2022_writer}. For the WV network, an architecture based on SigNet \cite{dey_2017_signet} with the selected ResNets as backbones is used. The networks are trained with the Triplet Loss \cite{keglevic_2018_learning} with a margin of 0.3 for the WI networks and the Contrastive Loss \cite{dey_2017_signet} for the WV networks. Both use the Cosine Similarity as distance measurement. The networks are trained with the Adam optimizer using a learning rate of 0.01 for WI and 0.001 for WV. 4-fold cross-validation is applied and the model with the highest Mean Average Precision (mAP) for WI and accuracy for WV is used. During training, random 400px $\times$ 400px snippets, which contain at least 2\% black pixels, are taken from the pages of the dataset. For ICDAR2013, the snippet size is set to 200px. The performances of the networks are shown in Table \ref{tab:eval_test_accuracy_nn}.
\begin{table}
\caption{Performances of the networks on the selected test sets. Values are given in percent.}
    \centering
  \begin{tabular}{cccc}
    \toprule
      && mAP & Accuracy  \\
    \midrule
    \multirow{3}{*}{CVL} & ResNet18  & \textbf{86.41}  & \textbf{90.48}     \\
    & ResNet20   & 56.43  & 67.58\\
    & ResNet50 & 82.41  & 88.52\\
    \midrule
    \multirow{3}{*}{Firemaker} & ResNet18 & \textbf{70.73}  & \textbf{93.4}     \\
    & ResNet20   & 39.66  & 68.4 \\
    & ResNet50  & 57.64  & 86.0  \\
    \midrule
    \multirow{3}{*}{ICDAR2013}  & ResNet18  & \textbf{62.08}  & \textbf{91.57}   \\
    & ResNet20   & 47.10 & 76.14\\
    & ResNet50   & 53.98 & 90.43\\
    \bottomrule
  \end{tabular}
  \label{tab:eval_test_accuracy_nn}
\end{table}

\subsection{Metrics}
\label{sec:metrics}
\begin{figure*}[tp!]
  \centering
  \hspace{0.4cm}
    \includegraphics[width=0.25\textwidth]{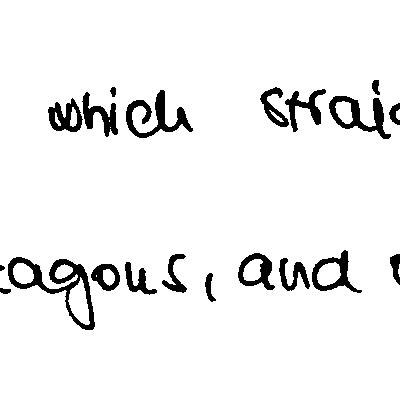}
    \hspace{3.8 cm}
    \includegraphics[width=0.25\textwidth]{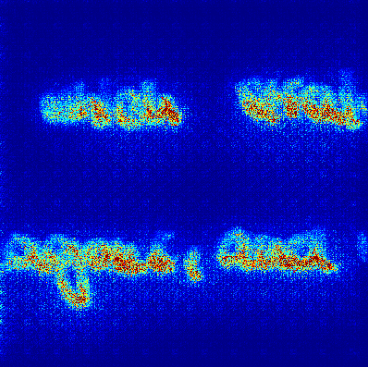}
    \vspace{0.3cm}
    \\
    \includegraphics[height=0.35\textwidth]{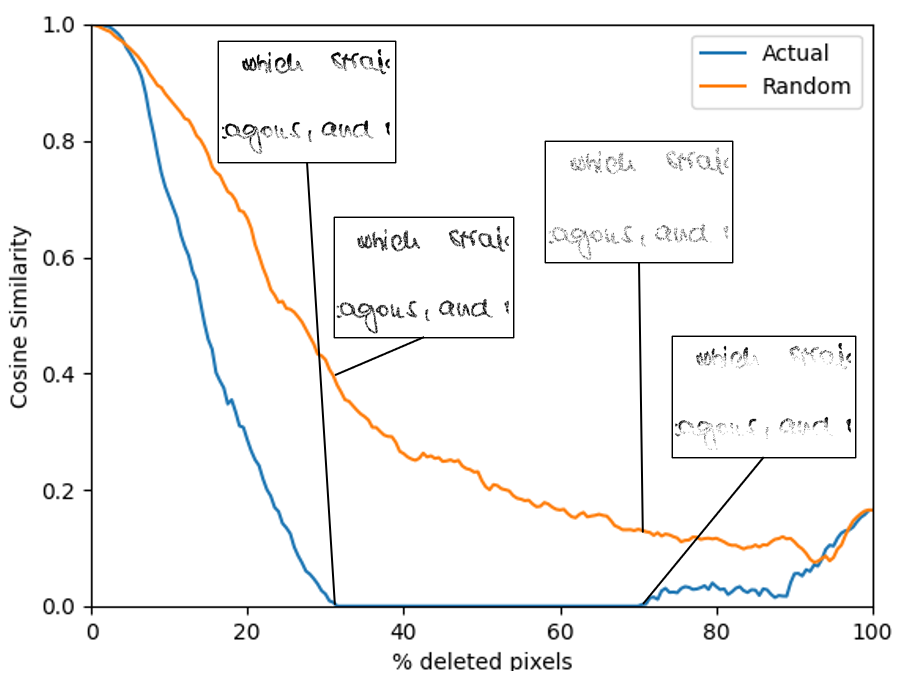}
    \includegraphics[height=0.345\textwidth]{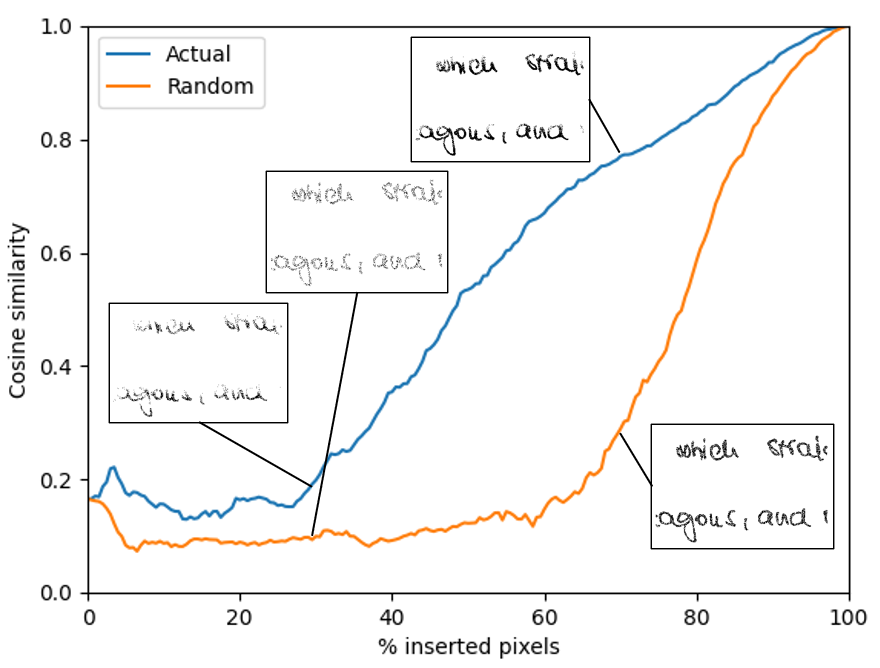}
  \caption{Example of deletion and insertion scores calculated for the given saliency map.} 
  \label{fig:examples_deletion_insertion_curves} 
\end{figure*}

For the evaluation of the transparency techniques, the deletion and insertion scores as described by Hu et al. \cite{hu_2022_xmir} are used. These metrics measure the change in similarity between a query and a retrieved image when the retrieved image is altered based on its given saliency map \cite{hu_2022_xmir}. For this work, the query and retrieved image are the same. Iteratively, the pixels of the retrieved image are altered based on the saliency map. For the deletion score, the most significant pixels are deleted first. For the insertion score, the most significant pixels are inserted first. The similarity between the images is then calculated using the Cosine Similarity. The value is set to zero if it falls below this threshold. The Area-Under-the-Curve (AUC) value is then calculated for the generated curve. For the evaluation, the deletion and insertion scores for the generated saliency maps are compared with the scores for a random deletion and insertion of pixels. The deletion score is calculated as $auc_d = \sum_{k=1}^{n} \frac{d(k)}{n} $ with
\begin{equation}
    d(k)=
    \begin{cases}
      1, & \text{if } r_k > s_k \\
      0, & \text{otherwise}
    \end{cases}
  \end{equation}
where $r_k$ is the AUC for the random deletion and $s_k$ is the AUC for the deletion according to the saliency map. The insertion score is calculated as $auc_i = \sum_{k=1}^{n} \frac{1-d(k)}{n}$. The deletion and insertion scores only affect black pixels, i.e. pixels, which are part of the handwriting. Pixels deleted during the calculation of the score are set to white, which simulates a progressive removal of handwriting information from the image. For the insertion score, a white image is used as the base, to which the pixels of the handwriting are gradually added. An example is shown in Figure \ref{fig:examples_deletion_insertion_curves}. The blue curve displays the change in similarity for the actual saliency map, while the orange curve displays the change in similarity for the random deletion and insertion of pixels. In this case, the curves indicate a correct highlighting provided by the saliency map, as the similarity between the original and altered image drops quicker than for the random deletion and increases quicker for the insertion.

\subsection{Datasets}
Three offline datasets containing contemporary handwriting are used for the training of the networks and the evaluation of the transparency techniques. The CVL dataset \cite{kleber_2013_cvl} contains 1604 handwritten pages, while the Firemaker \cite{firemaker} and ICDAR2013 \cite{louloudis_2013_icdar} datasets consist of 1000 pages each. For this work, the datasets are split into two open sets, therefore, the samples of one writer are assigned to one of the two sets. Both sets have 50\% of the amount of unique writers to have more training data available. The first set is used as train and validation set. The second set is used as test set.

\subsection{Quantitative evaluation}
The networks for WI and WV are trained with the CVL \cite{kleber_2013_cvl}, Firemaker \cite{firemaker} and ICDAR2013 \cite{louloudis_2013_icdar} datasets using 4-fold cross-validation. For the evaluation, the model with the highest test mAP in the case of WI neural networks and the highest accuracy in the case of WV neural networks is used. The pages of the test sets are divided into grids with snippets of 400px $\times$ 400px. The results of the quantitative evaluation for the deletion and insertion score calculations are shown in Table \ref{tab:eval_results_ins_del_wi}. For the transparency technique proposed by Zhu et al. \cite{zhu_2021_visual}, only the overall saliency maps are used for the quantitative evaluation as a manual point selection is infeasible for the point-specific saliency maps. 
\begin{table}
   \caption{$auc_d$ and $auc_i$ scores for the WI networks and the SigNet WV networks with different backbones.}
   \centering
   \begin{tabular}{cccccccccc}
   \toprule
   && \multicolumn{4}{c}{WI} &\multicolumn{4}{c}{WV} \\
   \cmidrule(r){3-10}
    && \multicolumn{2}{c}{Pixel-Level} & \multicolumn{2}{c}{Point-Specific} &\multicolumn{2}{c}{Pixel-Level} & \multicolumn{2}{c}{Point-Specific}\\
    \cmidrule(r){3-10}
    && $auc_d$ & $auc_i$ & $auc_d$ & $auc_i$ & $auc_d$ & $auc_i$ & $auc_d$ & $auc_i$ \\
   \midrule
   \multirow{3}{*}{CVL} &ResNet18  & \textbf{89.7} & 95.3 & \textbf{25.6} & 97.7 & \textbf{67.4} & 92.3 & 18.1 &97.1   \\ 
     &ResNet20   & 12.9 & 95.0 & 8.6 & \textbf{100.0} & 11.3 & 84.3 & \textbf{34.5} & \textbf{100.0}\\
     &ResNet50 & 55.6 & \textbf{98.3} & 8.3 & 98.8 & 61.1 & \textbf{96.7} & 28.6 & 89.8\\
   \midrule
      \multirow{3}{*}{Firemaker} &ResNet18 & \textbf{97.2} & \textbf{100.0} & 8.3 & \textbf{99.9} & 57.5 & 92.2 &8.4 &99.4    \\
    & ResNet20   & 15.6 & 95.7 & \textbf{22.8} & 99.8 &43.6 & \textbf{98.9} & \textbf{15.3} & \textbf{100.0}\\
     &ResNet50  & 51.3 & 89.3 & 21.4 & 99.3 & \textbf{69.2} & 97.7 & 13.9 & 99.0\\
\midrule
     \multirow{3}{*}{ICDAR2013} & ResNet18  & 82.3 & \textbf{99.9} & \textbf{8.0} & 99.6 & \textbf{75.6} & \textbf{99.7} &11.6 &99.6   \\
    &ResNet20   &15.1 & 99.6  &7.2 & \textbf{100.0} &6.2 & 98.3 & 10.3 & \textbf{100.0}\\
     &ResNet50   &\textbf{72.0} & 99.7 &  7.3 & 99.8 & 51.1 & 95.5 & \textbf{16.3}& 99.8\\
   \bottomrule
   \end{tabular}
   \label{tab:eval_results_ins_del_wi}
\end{table}
\paragraph{WI networks} The deletion scores of all three networks indicate a dependence of the performance of the pixel-level saliency maps on the underlying network architecture. The insertion scores for both types of saliency maps are high for the WI networks, with all but one value being above 90\%. This suggests that the networks are able to correctly classify a snippet of handwritten text with little information, i.e. when only a few pixels are inserted into the image. The results also show that the pixel-level saliency maps perform better than the point-specific saliency
maps regarding the deletion score for the WI neural networks. One possible explanation is that the deletion of individual pixels alters important handwriting information more drastically than the deletion of multiple pixels from one area of text, where the network can use unaltered areas with less significance within the snippet to identify the author.

\paragraph{WV networks} All insertion score percentages except two are above 90\% and all deletion score percentages except one fall below the threshold of 70\%. This also indicates that the networks are able to classify a snippet of handwritten text with only little information available. In contrast to the WI networks, the performance of the pixel-level saliency maps on the WV networks does not display a pattern regarding the underlying network. The deletion score percentage for ResNet20 is the lowest for all three datasets. However, the insertion score percentage achieved by ResNet20 is the highest for the Firemaker and ICDAR2013 datasets. The differences in the results for the WI and WV networks indicate an influence of the loss function and training procedure on the creation of the saliency maps.

\subsection{Qualitative evaluation}
\begin{figure}[htbp]
  \centering
   \begin{subfigure}[b]{0.48\columnwidth}
    \centering
    \includegraphics[width=0.9\textwidth]{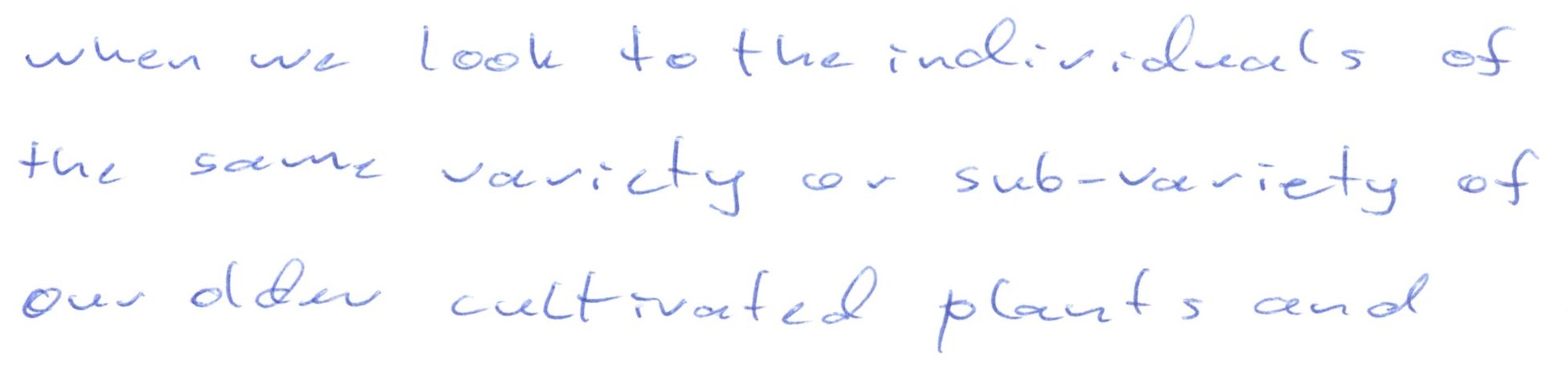}
    \subcaption{}
    \label{fig:original_snippets_0363-4}
  \end{subfigure}
  \begin{subfigure}[b]{0.48\columnwidth}
    \centering
    \includegraphics[width=0.9\textwidth]{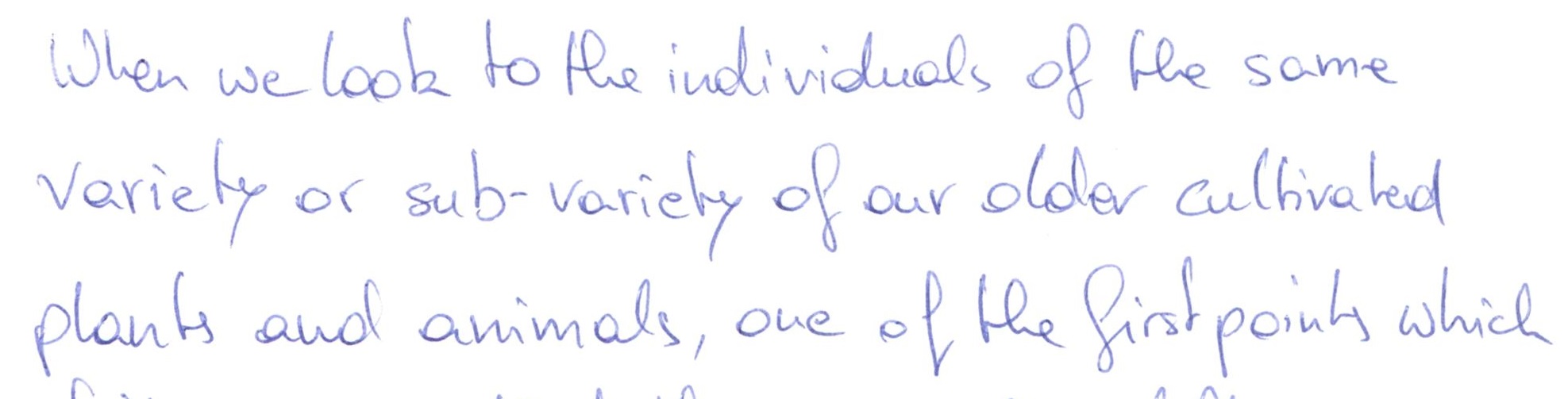}
    \subcaption{}
    \label{fig:original_snippets_0369-4}
  \end{subfigure}
  \caption{The first three lines of the pages taken from the test set of the CVL dataset \cite{kleber_2013_cvl} for the evaluation of the transparency techniques.}
  \label{fig:original_snippets} 
\end{figure}
For the qualitative evaluation, the saliency maps are analysed on similarities in highlightings of the same character occurrences in the same and different image snippets, since Christlein \cite{christlein_2019_handwriting} notes that the identification process done by a human expert can include the comparison of occurrences of the same character at different positions in the handwritten text. Due to the normalization of the values in the map by the transparency techniques, the intensity of the highlights are only meaningful in the scope of its snippet, as at least one area with the highest possible significance highlight exists in each snippet. Therefore, the highlights do not provide information on absolute values but display a relative significance between areas of the same snippet.
\begin{table}
   \caption{Performance of the first and second page.}
   \centering
   %\begin{subtable}[t]{0.48\textwidth}
   %\centering
   \begin{tabular}{ccccc}
   \toprule
    & \multicolumn{2}{c}{WI} & \multicolumn{2}{c}{WV}\\
    & \multicolumn{2}{c}{Average Precision} & \multicolumn{2}{c}{ Matches}   \\
     &First & Second & First & Second\\
   \midrule
    ResNet18 & 1 & 0.04 & 100\% & 0\%  \\ 
     ResNet20  & 0.92  & 0.02 & 100\% & 75\%   \\
     ResNet50  & 1 & 0.04 & 100\% & 0\%   \\
   \bottomrule
   \end{tabular}
   \label{tab:performance_pages}
\end{table}

Two images from different authors are selected from the test set of the CVL dataset for evaluation. The selected pages are shown in Figures \ref{fig:original_snippets_0363-4} and \ref{fig:original_snippets_0369-4}. For the first page, the retrieval for the WI networks works, while for the second page, the retrieval does not work. The according average precision values are shown in Table \ref{tab:performance_pages}. For the WV networks, the matching of the first page is correct, while for the second page, the matching is incorrect. The percentage of pages written by the same writer, which are returned as matches, is shown in Table  \ref{tab:performance_pages} as well. The neural networks have been trained with the CVL dataset beforehand. The saliency maps for the selected images are calculated for ResNet18 and ResNet20, since the saliency maps for ResNet50 are similar to the ones for ResNet18.

\begin{figure}[htbp]
  \centering
  \begin{subfigure}[b]{0.24\columnwidth}
    \centering
    \includegraphics[width=0.9\textwidth]{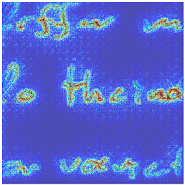}
    \caption{WI ResNet18}
  \end{subfigure}
    \begin{subfigure}[b]{0.24\columnwidth}
        \centering
        \includegraphics[width=0.9\textwidth]{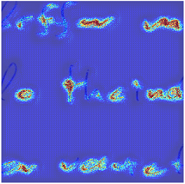}
        \caption{WI ResNet20}
    \end{subfigure}
     \begin{subfigure}[b]{0.24\columnwidth}
    \centering
    \includegraphics[width=0.9\textwidth]{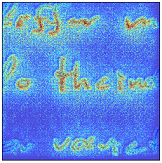}
    \caption{WV ResNet18}
  \end{subfigure}
   \begin{subfigure}[b]{0.24\columnwidth}
    \centering
    \includegraphics[width=0.9\textwidth]{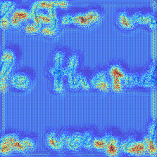}
    \caption{WV ResNet20}
  \end{subfigure}
  \caption{Examples for the pixel-wise saliency maps for the handwritten text of the first page.} 
  \label{fig:CVL_WI_pw_pages} 
\end{figure}

\begin{figure}[h]
  \centering
  \begin{subfigure}[b]{0.24\columnwidth}
    \centering
    \includegraphics[height=0.34\textwidth]{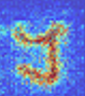}
    \includegraphics[height=0.34\textwidth]{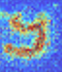}
    \includegraphics[height=0.34\textwidth]{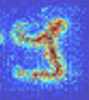}
    \caption{ResNet18 First Page}
  \end{subfigure}
  \begin{subfigure}[b]{0.24\columnwidth}
    \centering
    \includegraphics[height=0.34\textwidth]{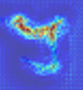}
    \includegraphics[height=0.34\textwidth]{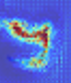}
    \includegraphics[height=0.34\textwidth]{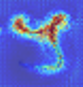}
    \caption{ResNet20 First Page}
  \end{subfigure}
  \begin{subfigure}[b]{0.24\columnwidth}
    \centering
    \includegraphics[height=0.35\textwidth]{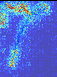}
    \includegraphics[height=0.35\textwidth]{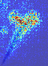}
    \includegraphics[height=0.35\textwidth]{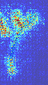}
    \caption{ResNet18 Second Page}
  \end{subfigure}
  \begin{subfigure}[b]{0.24\columnwidth}
    \centering
    \includegraphics[height=0.35\textwidth]{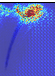}
    \includegraphics[height=0.35\textwidth]{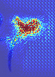}
    \includegraphics[height=0.35\textwidth]{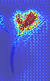}
    \caption{ResNet20 Second Page}
  \end{subfigure}
  \caption{Reoccurring highlighting patterns for the occurrences of the character ''y''.} 
  \label{fig:y_pw_wi} 
\end{figure}

\begin{figure}[h]
  \centering
  \begin{subfigure}[b]{0.24\columnwidth}
    \centering
    \includegraphics[height=0.4\textwidth]{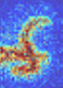}
    \includegraphics[height=0.4\textwidth]{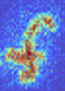}
    \includegraphics[height=0.4\textwidth]{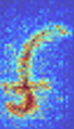}
    \caption{ResNet18 First Page}
  \end{subfigure}
  \begin{subfigure}[b]{0.24\columnwidth}
    \centering
    \includegraphics[height=0.4\textwidth]{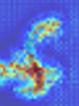}
    \includegraphics[height=0.4\textwidth]{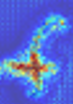}
    \includegraphics[height=0.4\textwidth]{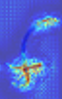}
    \caption{ResNet20 First Page}
  \end{subfigure}
   \begin{subfigure}[b]{0.24\columnwidth}
    \centering
    \includegraphics[height=0.4\textwidth]{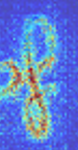}
    \includegraphics[height=0.4\textwidth]{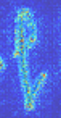}
    \includegraphics[height=0.4\textwidth]{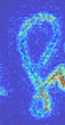}
    \caption{ResNet18 Second Page}
  \end{subfigure}
  \begin{subfigure}[b]{0.24\columnwidth}
    \centering
    \includegraphics[height=0.4\textwidth]{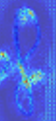}
    \includegraphics[height=0.4\textwidth]{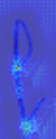}
    \includegraphics[height=0.4\textwidth]{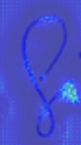}
    \caption{ResNet20 Second Page}
  \end{subfigure}
  \caption{Occurrences of the character ''f''. The highlights for the first page display a pattern, while no pattern emerges for the second page.} 
  \label{fig:f_pw_wi} 
\end{figure}

\paragraph{Pixel-Wise Saliency Map}
Examples for the pixel-wise saliency maps generated for the WI and WV networks for the first page are shown in Figure \ref{fig:CVL_WI_pw_pages}.
The saliency maps for the first page display similar highlighting patterns for multiple occurrences of the character ''y'', as shown in Figure \ref{fig:y_pw_wi}, and ''f'', as shown in Figure \ref{fig:f_pw_wi}. The highlighting pattern differs from network to network but stays consistent for the saliency maps created for one neural network type. However, both networks also display deviating highlightings for their occurrences of characters. %such as ''p'', as shown in Figure \ref{fig:p_pw_wi_18_0363}. 

In contrast to the first page, the saliency maps for the second page display large differences in their highlighting patterns. For example, the character ''f'' does not display a similar highlighting, which is shown in Figure \ref{fig:f_pw_wi}.
\begin{figure}[h]
  \centering
  \begin{subfigure}[b]{0.24\columnwidth}
    \centering
    \includegraphics[width=0.9\textwidth]{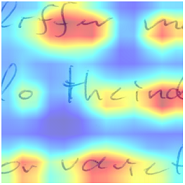}
    \caption{WI ResNet18}
    \label{fig:CVL_WI_ps_page_a}
  \end{subfigure}
  \begin{subfigure}[b]{0.24\columnwidth}
    \centering
    \includegraphics[width=0.9\textwidth]{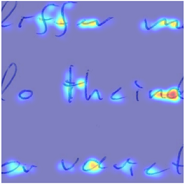}
    \caption{WI ResNet20}
  \end{subfigure}
   \begin{subfigure}[b]{0.24\columnwidth}
    \centering
    \includegraphics[width=0.9\textwidth]{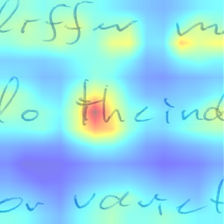}
    \caption{WV ResNet18}
  \end{subfigure}
  \begin{subfigure}[b]{0.24\columnwidth}
    \centering
    \includegraphics[width=0.9\textwidth]{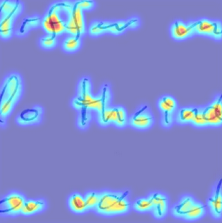}
    \caption{WV ResNet20}
  \end{subfigure}
  \caption{Examples for the point-specific saliency maps for the handwritten text of the first page.} 
  \label{fig:CVL_WI_ps_page} 
\end{figure}
%A comparison of the saliency maps for the WI and WV networks shows that differing highlighting patterns are displayed. Examples are shown in Figure \ref{fig:f_pw_wi_vs_wv}.
%\begin{figure}[tp!]
%  \centering
%  \begin{subfigure}[b]{0.3\columnwidth}
%    \centering
%    \includegraphics[height=0.32\textwidth]{images/character_snippets/f_pw_wi_18_0363_1.PNG}
%    \includegraphics[height=0.32\textwidth]{images/character_snippets/f_pw_wi_18_0363_2.PNG}
%    \includegraphics[height=0.32\textwidth]{images/character_snippets/f_pw_wi_18_0363_3.PNG}
%    \caption{WI ResNet18}
%  \end{subfigure}
%  \begin{subfigure}[b]{0.3\columnwidth}
%    \centering
%    \includegraphics[height=0.32\textwidth]{images/character_snippets/f_pw_wv_18_0363_1.PNG}
%    \includegraphics[height=0.32\textwidth]{images/character_snippets/f_pw_wv_18_0363_2.PNG}
%    \includegraphics[height=0.32\textwidth]{images/character_snippets/f_pw_wv_18_0363_3.PNG}
%    \caption{WV ResNet18}
%  \end{subfigure}
%  \caption{Occurrences of the character ''f'', where a highlighting pattern is visible.} 
%  \label{fig:f_pw_wi_vs_wv} 
%\end{figure}

A comparison of the saliency maps for the first and second page shows that the maps for the snippets of the first page contain more secluded high intensity areas in comparison to the second page, which displays multiple areas and characters with high intensity. This indicates that the snippets of the second page do not have salient characteristics, which the neural networks can use to identify the author and all available information is used to determine the output of the neural network.

\paragraph{Point-Specific Saliency Map}

Examples for the point-specific saliency maps generated for the WI and WV networks for the first page are shown in Figure \ref{fig:CVL_WI_ps_page}. 

\begin{figure}[tp!]
  \centering
   \begin{subfigure}[b]{0.24\columnwidth}
    \centering
    \includegraphics[height=0.39\textwidth]{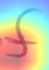}
    \includegraphics[height=0.39\textwidth]{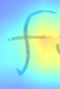}
    \includegraphics[height=0.39\textwidth]{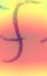}
    \caption{ResNet18 First Page}
  \end{subfigure}
  \begin{subfigure}[b]{0.24\columnwidth}
    \centering
    \includegraphics[height=0.39\textwidth]{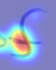}
    \includegraphics[height=0.39\textwidth]{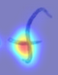}
    \includegraphics[height=0.39\textwidth]{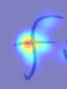}
    \caption{ResNet20 First Page}
  \end{subfigure}
  \begin{subfigure}[b]{0.24\columnwidth}
    \centering
    \includegraphics[height=0.39\textwidth]{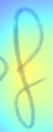}
    \includegraphics[height=0.39\textwidth]{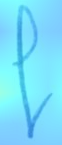}
    \includegraphics[height=0.39\textwidth]{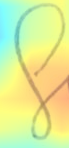}
    \caption{ResNet18 Second Page}
  \end{subfigure}
    \begin{subfigure}[b]{0.24\columnwidth}
    \centering
    \includegraphics[height=0.39\textwidth]{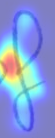}
    \includegraphics[height=0.39\textwidth]{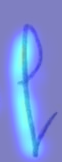}
    \includegraphics[height=0.39\textwidth]{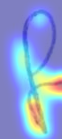}
    \caption{ResNet20 Second Page}
  \end{subfigure}
  \caption{Occurrences of the character ''f''. A highlighting pattern is visible for ResNet20 for the first page, while no patterns emerges for the second page and for both pages of ResNet18.} 
  \label{fig:f_occurrences_ps_wi} 
\end{figure}

The saliency maps for ResNet20 display a similar highlighting pattern for the character ''f'' for the first page. However, for the second page, no highlighting pattern emerges. Additionally, no highlighting pattern emerges for ResNet18 for both pages. Examples are shown in Figure \ref{fig:f_occurrences_ps_wi}.

A comparison of the saliency maps for the WI and WV networks shows that differing highlighting patterns are displayed as well. ResNet18 does not display a reoccurring pattern. ResNet20 displays a reoccurring pattern, where one end of the lines, which constructs the character, is highlighted.

The allocation of a highlight to a certain character is aggravated for ResNet18, as it covers large areas of the page and, in some cases, overlaps multiple characters. In contrast, ResNet20 displays concise highlighting areas, which can be assigned to a character. This shows the influence of the feature map size of the last convolutional layer on the accuracy of the highlighting regarding its allocation, as the feature maps for ResNet20 are significantly larger than the feature maps for ResNet18.

\paragraph{Point-to-Image Comparison}
The point-specific saliency map technique provides a second type of saliency map, where a point in an image is selected for a similarity analysis with a second image. In this section, the analysis is conducted between a point in an image snippet and a second snippet taken from another page written by the same author. A ResNet18 model is used as neural network. 

\begin{figure}[htbp]
  \centering
  \begin{subfigure}[b]{0.24\columnwidth}
    \centering
    \includegraphics[width=0.9\textwidth]{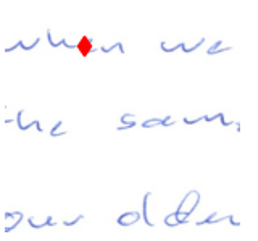}
    \subcaption{}
  \end{subfigure}
 \begin{subfigure}[b]{0.24\columnwidth}
    \centering
    \includegraphics[width=0.9\textwidth]{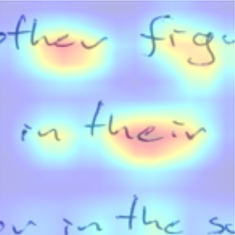}
    \subcaption{}
  \end{subfigure}
  \begin{subfigure}[b]{0.24\columnwidth}
    \centering
    \includegraphics[width=0.9\textwidth]{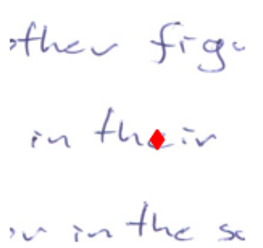}
    \subcaption{}
  \end{subfigure}
  \begin{subfigure}[b]{0.24\columnwidth}
    \centering
    \includegraphics[width=0.9\textwidth]{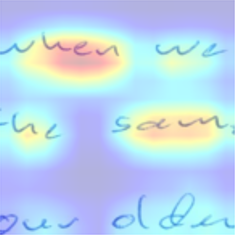}
    \subcaption{}
  \end{subfigure}
  \caption{Point-specific saliency maps for the first author with a selected point. The selected points are displayed as red diamonds.}
  \label{fig:ps_eval_point_0363} 
\end{figure}

\begin{figure}[htbp]
  \centering
  \begin{subfigure}[b]{0.05\columnwidth}
    \centering
    \includegraphics[width=0.9\textwidth]{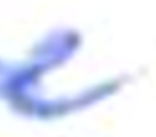}
    \subcaption{}
  \end{subfigure}
 \begin{subfigure}[b]{0.05\columnwidth}
    \centering
    \includegraphics[width=0.9\textwidth]{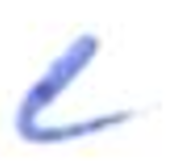}
    \subcaption{}
  \end{subfigure}
  \caption{Character parts taken from the page shown in Figure \ref{fig:original_snippets_0363-4}. The left snippet displays the right part of the character ''a'' while the right snippet displays an occurrence of the character ''e''.}
  \label{fig:e_a_comparison} 
\end{figure}

\begin{figure}[h!]
  \centering
  \begin{subfigure}[b]{0.24\columnwidth}
    \centering
    \includegraphics[width=0.9\textwidth]{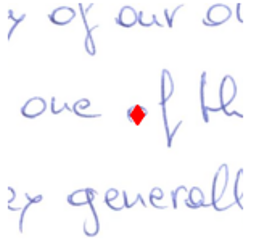}
    \subcaption{}
  \end{subfigure}
 \begin{subfigure}[b]{0.24\columnwidth}
    \centering
    \includegraphics[width=0.9\textwidth]{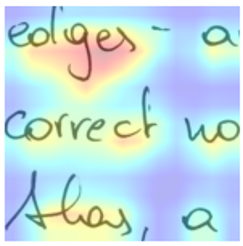}
    \subcaption{}
  \end{subfigure}
  \begin{subfigure}[b]{0.24\columnwidth}
    \centering
    \includegraphics[width=0.9\textwidth]{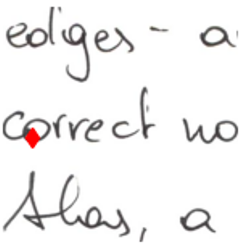}
    \subcaption{}
  \end{subfigure}
  \begin{subfigure}[b]{0.24\columnwidth}
    \centering
    \includegraphics[width=0.9\textwidth]{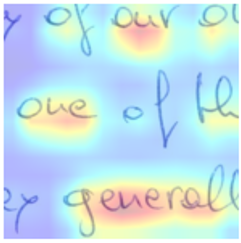}
    \subcaption{}
  \end{subfigure}
  \caption{Point-specific saliency maps for the second author with a selected point. The selected points are displayed as red diamonds.}
  \label{fig:ps_eval_point_0369} 
\end{figure}

The saliency maps generated for the snippets taken from two pages written by the first author are shown in Figure \ref{fig:ps_eval_point_0363}. For these snippets, the three highlights with the highest significance are not consistently placed on occurrences of the same character. For the second snippet, one of the highlights is placed on the character ''a'', where the right part of the character ''a'', resembles the characteristics of the character ''e''. This similarity is shown in Figure \ref{fig:e_a_comparison}.
The saliency maps generated for the snippets taken from two pages written by the second author are shown in Figure \ref{fig:ps_eval_point_0369}. 
In both cases, the actual occurrences of the selected character are not highlighted. Instead, the highlight is placed on characters with a similar round bottom part. However, this highlight is not placed consistently.

\section{Conclusion}
In this work, the applicability of transparency techniques on WI and WV neural networks is evaluated. The goal is to gain insight into the decision process of the underlying network and support investigators with information on similarities in the given handwritten text. For this purpose, two transparency techniques, namely the pixel-wise saliency maps proposed by Kobs et al.~\cite{kobs_2021_different} and the point-specific saliency maps proposed by Zhu et al.~\cite{zhu_2021_visual}, are selected.

The quantitative evaluation results show, that the pixel-level saliency maps perform better than the point-specific saliency maps for the deletion score. The difference in performance indicates that the pixel-wise saliency maps provide more accurately allocated highlightings than the point-specific saliency maps. For the qualitative evaluation results, the pixel-wise saliency maps frequently display similar highlightings for multiple character occurrences. This indicates that the networks analyse these occurrences similar to a human investigator. The point-specific saliency maps, however, do not display similar highlightings with reliable frequency. Additionally, the position of a peak highlight is difficult to allocate to a certain character. The networks consider few characters with high significance if multiple characters and, therefore, more information is available, indicating a focus on salient characteristics. The upscaling of the calculated values for point-specific saliency maps indicates that this transparency technique performs well with little information in concise locations. This suggests that the use of the point-specific saliency maps with character-based input images, i.e. images containing only one handwritten character, could improve the level of detail the saliency maps can provide.

Overall, the similar highlightings for different occurrences of the same character for the pixel-wise saliency maps shows that this transparency technique can support the analysis of a handwritten text by an investigator. However, the point-specific saliency maps display non-intuitive highlightings regarding the allocation of a highlight to a character and are not suitable for the support of the analysis process.

Future work could include an evaluation of the similarity of highlights for a character-based neural network, i.e. a network which takes input images containing a single character, where the intensity of the highlight can be taken into account.

\medskip
{
\small

\bibliography{sources}

%[1] Alexander, J.A.\ \& Mozer, M.C.\ (1995) Template-based algorithms for
%connectionist rule extraction. In G.\ Tesauro, D.S.\ Touretzky and T.K.\ Leen
%(eds.), {\it Advances in Neural Information Processing Systems 7},
%pp.\ 609--616. Cambridge, MA: MIT Press.

%[2] Bower, J.M.\ \& Beeman, D.\ (1995) {\it The Book of GENESIS: Exploring
%  Realistic Neural Models with the GEneral NEural SImulation System.}  New York:
%TELOS/Springer--Verlag.

%[3] Hasselmo, M.E., Schnell, E.\ \& Barkai, E.\ (1995) Dynamics of learning and
%recall at excitatory recurrent synapses and cholinergic modulation in rat
%hippocampal region CA3. {\it Journal of Neuroscience} {\bf 15}(7):5249-5262.
}

%%%%%%%%%%%%%%%%%%%%%%%%%%%%%%%%%%%%%%%%%%%%%%%%%%%%%%%%%%%%

\end{document}